# RoLA: A Real-Time Online Lightweight Anomaly Detection System for Multivariate Time Series

Ming-Chang Lee[1] and Jia-Chun Lin[2]

[1]Department of Computer science, Electrical engineering and Mathematical sciences, Høgskulen på Vestlandet (HVL), Bergen, Norway

[2]Department of Information Security and Communication Technology, Norwegian University of Science and Technology, Gjøvik, Norway

[1] mingchang1109@gmail.com
[2] jia-chun.lin@ntnu.no

26th May 2023

# RoLA: A Real-Time Online Lightweight Anomaly Detection System for Multivariate Time Series

Ming-Chang Lee[1] [a] and Jia-Chun Lin[2] [b]

[1]*Department of Computer science, Electrical Engineering and Mathematical Sciences, Høgskulen på Vestlandet (HVL), Bergen, Norway*
[2]*Department of Information Security and Communication Technology, Norwegian University of Science and Technology (NTNU), Gjøvik, Norway*
mingchang1109@gmail.com, jia-chun.lin@ntnu.no

Abstract: A multivariate time series refers to observations of two or more variables taken from a device or a system simultaneously over time. There is an increasing need to monitor multivariate time series and detect anomalies in real time to ensure proper system operation and good service quality. It is also highly desirable to have a lightweight anomaly detection system that considers correlations between different variables, adapts to changes in the pattern of the multivariate time series, offers immediate responses, and provides supportive information regarding detection results based on unsupervised learning and online model training. In the past decade, many multivariate time series anomaly detection approaches have been introduced. However, they are unable to offer all the above-mentioned features. In this paper, we propose RoLA, a <u>r</u>eal-time <u>o</u>nline <u>l</u>ightweight <u>a</u>nomaly detection system for multivariate time series based on a divide-and-conquer strategy, parallel processing, and the majority rule. RoLA employs multiple lightweight anomaly detectors to monitor multivariate time series in parallel, determine the correlations between variables dynamically on the fly, and then jointly detect anomalies based on the majority rule in real time. To demonstrate the performance of RoLA, we conducted an experiment based on a public dataset provided by the FerryBox of the One Ocean Expedition. The results show that RoLA provides satisfactory detection accuracy and lightweight performance.

## 1 INTRODUCTION

A multivariate time series consists of sequences of values of several simultaneous variables changing with time (Chakraborty et al., 1992). In the real world, multivariate time series are continuously generated by sensors of industry devices or large systems such as server machines, spacecrafts, engines, water treatment plant, power grids, etc (Su et al., 2019).

Monitoring multivariate time series and detecting anomalies in the time series has become an imperative task for critical infrastructures such as transportation systems, communication networks, and diverse cyber-physical systems. There is an increasing need to have a real-time, reliable, and accurate anomaly detection approach that does not rely on labeled data or supervised learning and that is fast enough to provide instant reporting (Wu et al., 2020). It is also essential to take the correlations between different variables of a multivariate time series into consideration in order to reduce false positives (Zhao et al., 2020). In addition, it would be desirable to have a lightweight anomaly detection approach that can adapt to changes in the pattern of the time series and requires no excessive computation resources (Lee et al., 2020b; Lee et al., 2021; Lee and Lin, 2023).

A number of multivariate time series anomaly detection approaches have previously been introduced based on machine learning. Many of them are based on either supervised or semi-supervised learning, which was found ineffective and infeasible in real-world applications. Some approaches have a complex design and require substantial human intervention to tune and configure different hyperparameters or parameters. Most of the approaches rely on offline model training, and therefore unable to adapt to changes in multivariate time series over time. In addition, almost all approaches act as a black box without explicitly learning the relationship between variables, which therefore limits their detection ability and disable them from providing interpretation when anomalies occur (Pang and Aggarwal, 2021; Deng and Hooi, 2021).

[a]  https://orcid.org/0000-0003-2484-4366
[b]  https://orcid.org/0000-0003-3374-8536

In this paper, we propose a real-time online lightweight anomaly detection system (RoLA for short) for multivariate time series following a divide-and-conquer strategy, parallel processing, and the majority rule. RoLA employs a set of lightweight anomaly detectors (LADs for short) to detect anomalies in multivariate time series in parallel based on RePAD2 (Lee and Lin, 2023), which is a state-of-the-art real-time and lightweight anomaly detection approach for open-ended univariate time series.

Whenever a LAD detects a suspicious data point for the variable it has been tracking, RoLA will calculate the correlation between that variable and every other variable by tracing back their time series in a recent period based on the Pearson correlation coefficient (Cohen et al., 2009). After that, all highly-correlated variables will go through a polling process. If more than half of them have a suspicious data point detected by their own LADs at that moment, RoLA will report that all these data points are anomalous and output all relevant variables as supportive information. It is worth noting that RoLA requires no offline model training, labeled data, or pre-defined detection threshold. Its simple and lightweight design makes itself to be easily deployed on any commodity computer.

To demonstrate the performance of RoLA in terms of detection accuracy, time consumption, and instant response, we conducted an experiment based on real-world multivariate time series generated by FerryBox (Schroeder et al., 2008) of the One Ocean Expedition (King et al., 2021; Knoblauch, 2022) by ingesting the multivariate time series into RoLA via Apache Kafka (Garg, 2013) to mimic the real-world streaming behavior. Note that Apache Kafka is an open-source distributed streaming platform developed by the Apache Software Foundation for real-time stream processing.

The results show that RoLA provides satisfactory detection accuracy and offers instant response to every detected anomaly without needing any general-purpose GPUs or high-performance computers. Furthermore, RoLA provides useful information about what variables are involved in each detected anomaly. Such information would be helpful for domain experts to further investigate the anomalies and identify the events that caused the anomalies. The contributions of RoLA are summarized as follows:

1. Unsupervised learning and online model training: RoLA does not require any offline model training, labelled dataset or normal dataset to learn a multivariate time series. Instead, RoLA individually learns and monitors each time series of a multivariate time series in an online manner. Furthermore, RoLA also learns the dynamic correlation between different variables whenever any LAD detects a suspicious data point. These features enable RoLA to capture dynamic and unforeseen correlations among all variables.

2. Adaptability: In RoLA, each LAD automatically and periodically updates its detection threshold and retrains its model when needed. Hence, RoLA can adapt to changes in the pattern of the multivariate time series over time. There is no need for users to determine any detection threshold.

3. Real-time processing, instant response, and supportive information: RoLA can process multivariate time series in real time and provide an instant response whenever an anomaly occurs. Each detected anomaly also comes with useful information about all relevant variables.

4. Lightweight design without much human intervention: The design of RoLA is simple and lightweight. There is no need for users to configure many hyperparameters or parameters. Only two variables are needed to be determined beforehand, which will be described later.

5. No additional data storage: By incorporating Apache Kafka into the design of RoLA, multivariate time series can be directly ingested into RoLA and processed by RoLA without requiring any additional large data storage.

The rest of the paper is organized as follows: Section 2 discusses related work. In Section 3, we introduce the details of RoLA. Section 4 presents and discusses the experiment and the corresponding results. In Section 5, we conclude this paper and outline future work.

## 2 RELATED WORK

Anomaly detection in multivariate time series has been an active research topic, and many related approaches have been introduced. Conventional supervised learning approaches rely on labeled data for model training and can only detect known anomalies. Due to lack of labeled data in many real-world applications, many anomaly detection approaches were design based on either semi-supervised learning or unsupervised learning. In semi-supervised anomaly detection, training data is assumed to include only normal data. However, the training data in unsupervised anomaly detection is assumed to include a small minority of abnormal data instances (Chandola et al., 2009).

Hundman et al. (Hundman et al., 2018) handle multivariate time series anomaly detection by creating a model to detect anomalies for each spacecraft telemetry variable based on the Long Short-Term Memory (LSTM) and introducing a dynamic thresholding method to determine the detection threshold at each time step. Even though this approach is unsupervised, it is based on offline model training. Individually training a model for each telemetry variable can indeed speed up the learning, but this approach did not take correlations between different variables into consideration. Therefore, it might not be able to detect anomalies that propagate to multiple variables.

Audibert et al. (Audibert et al., 2020) proposed an unsupervised anomaly detection approach called USAD for multivariate time series based on autoencoders and adversarial training. USAD needs to go through three stages of offline training (autoencoder training, adversarial training, and two-phase training) even though their experiment results show that their training time is short. Due to the offline training, USAD might not be able to adapt to changes or capture unforeseen patterns in multivariate time series over time.

Deng and Hooi (Deng and Hooi, 2021) proposed a graph neural network-based anomaly detection for multivariate time series by learning a graph of the dependence relationships between sensors, computing individual anomalousness scores for each sensor, and combining them into a single anomalousness score. Similar to many other anomaly detection methods, this approach still requires offline training to learn the relationships between sensors, meaning that their model will not be able to capture relationship changes between sensors over time.

Rettig et al. (Rettig et al., 2019) introduced an online anomaly detection approach for signaling traffic of a mobile cellular network based on relative entropy and the Pearson correlation coefficient. However, they require both normal dataset and anomalous dataset, and require domain knowledge to set detection thresholds. Yao et al. (Yao et al., 2010) proposed an online anomaly detection method for wireless sensor systems by constructing a piecewise linear model for multivariate time series and comparing the piecewise linear models of sensor data collected during a time interval with a reference model. If there are significant differences, the data is flagged as anomalies. Apparently that this approach depends on the reference model, which requires support from domain experts and it needs to be done in advance.

According to the comprehensive review on outlier/anomaly detection for time series (Blázquez-García et al., 2021), current research gaps include how to design dynamic and adaptive detection thresholds, take complex correlations between variables of multivariate time series into consideration, and develop real-time anomaly detection that can provide immediate responses in real time and can be practically used in real-world applications.

Laxhammar and Falkman (Laxhammar and Falkman, 2013) also pointed out that many methods have the following limitations: invalid statistical assumption, parameter-laden, ad-hoc anomaly thresholds, offline learning, and offline anomaly detection. In addition, a lot of anomaly detection methods are constructed as a black box without being able to provide any information or explanation to their detection results (Pang and Aggarwal, 2021), or they do not explicitly learn the relationships between different variables of multivariate time series, therefore unable to detect and explain deviations from such relationships when anomalous events happen. (Deng and Hooi, 2021)

Different from many existing multivariate time series anomaly detection approaches, RoLA is a novel real-time online lightweight anomaly detection system. Each variable of a multivariate time series is continuously learned and monitored by a lightweight anomaly detector, which retrains its detection model only when it is needed based on a dynamically calculated detection threshold. All lightweight anomaly detectors work together to detect anomalies based on the corrections between variables and the majority rule. In addition, RoLA is also capable of offering supportive information, which allows domain experts to further investigate any detected anomaly. Because of its lightweight design, RoLA can provide instant responses in real time without requiring general-purpose GPU or high-performance computers.

To our best knowledge, RoLA is the first approach that can learn multivariate time series in a completely online manner and provide real-time and transparent anomaly detection without requiring any labeled dataset, normal dataset, or reference model.

## 3 RoLA

Recall that RoLA is proposed to be a real-time, online and lightweight anomaly detection system for multivariate time series and to provide instant responses and supportive information without much human intervention. In order to achieve this ambitious goal, we follow a divide-and-conquer strategy, parallel processing, and the majority rule to design RoLA. Instead of using one complex model to learn the target multivariate time series, RoLA employs one lightweight anomaly detector (i.e., LAD) to learn and monitor each individual variable of the target multivariate time series, and all LADs work together to detect anomalies in the multivariate time series. Each LAD is based on RePAD2 (Lee and Lin, 2023), which is a state-of-the-art real-time and lightweight anomaly detection approach for open-ended univariate time series.

Just like RePAD2, LAD always uses three historical data points to predict each upcoming data point in the target time series based on LSTM, a neural network designed to learn long-term dependencies and model temporal sequences (Hochreiter and Schmidhuber, 1997). LAD inherits a simple LSTM structure (only one hidden layer and ten hidden units) from RePAD2 (Lee and Lin, 2023) and RePAD (Lee et al., 2020b). By dynamically calculating a detection threshold at each time point and retraining a new LSTM model only when the current LSTM model cannot predict the next data point value accurately, LAD can adapt to any unforeseen changes in the pattern of the time series on the fly. LAD only considers the data point at the current time point as anomalous when it has tried retraining a new LSTM mode but the model still results in an AARE value higher than the detection threshold. Note that AARE stands for average absolute relative error.

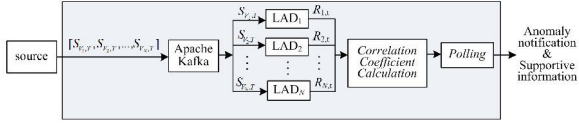

Figure 1: The architecture of RoLA.

Figures 1 and 2 illustrate the architecture and algorithm of RoLA, respectively. Suppose that the target multivariate time series consists of $N$ variables, denoted by $V_1$, $V_2$, ..., and $V_N$. Let $S_{V_x}$ be the time series of variable $V_x$, $x$=1, 2, ..., $N$. Assume that the source (which can be a device or a server) observes a new data point from every variable at every time point. Let $S_{V_x,T}$ be the data point of variable $V_x$ observed at $T$, where $T$ denotes the current time point and $T$ starts from 0. Whenever the source observes a N-dimensional vector $[S_{V_1,T}, S_{V_2,T}, ..., S_{V_N,T}]$ at $T$, this vector will be immediately sent to Apache Kafka. Hence, RoLA does not require any extra data storage to keep the time series. After that, Kafka distributes data points $S_{V_1,T}$, $S_{V_2,T}$, ..., and $S_{V_N,T}$ to $LAD_1$, $LAD_2$, ..., and $LAD_N$, respectively. Note that $LAD_x$ is responsible for learning time series $S_{V_x}$ and detecting anomalous data points in $S_{V_x}$.

Let $R_{x,T}$ be the detection result generated by $LAD_x$, and it indicates either data point $S_{V_x,T}$ is anomalous or not. As long as the current data point of a variable is detected as anomalous by its LAD, RoLA will calculate the Pearson correlation coefficient between that variable and every other variable in a recent period to find out which variables are highly correlated with that variable. Then all highly-correlated variables will go through a polling process.

**Input**: A multivariate time series of variables $V_1$, $V_2$, ..., and $V_N$
**Output**: Anomaly notification and supportive information
**Procedure:**
1: Let $A$, $L_{var}$, and $L_{data}$ be three empty lists;
2: Let $C_{agree}$ and $C_{disagree}$ be two counters with an initial value of 0;
3: **While** time has advanced {
4:     Let $T$ be the current time point;
5:     Receive $[S_{V_1,T}, S_{V_2,T}, ..., S_{V_N,T}]$ from the source;
6:     **for** $x$ =1 to $N$ { Use Kafka to distribute $S_{V_x,T}$ to $LAD_x$;}
7:     **for** $x$ =1 to $N$ {
8:         Let $R_{x,T}$ be the detection result generated by $LAD_x$;
9:         **if** $R_{x,T}$ shows that $S_{V_x,T}$ is anomalous {
10:             Add variable $V_x$ to $A$;}}
11:     **if** $|A| > 0$ {
12:         **for** $y$ = 1 to $|A|$ {
13:             Set $C_{agree}$ to one; reset $C_{disagree}$ to zero; Empty both $L_{var}$ and $L_{data}$;
14:             Let $a$ denote the $y$-th variable on $A$; Add $a$ to $L_{var}$;
15:             Let $S_{a,T}$ be the current data point of variable $a$; Add $S_{a,T}$ to $L_{data}$;
16:             **for** $z$ =1 to $N$ {
17:                 **if** $V_z$ is a different variable from $a$ {
18:                     Let $b$ denote variable $V_z$;
19:                     Let $E_{a,b}$ be the correlation coefficient between $a$ and $b$;
20:                     Calculate $E_{a,b}$ based on Equation 5;
21:                 **if** $E_{a,b} \geq thd_{pos}$ or $E_{a,b} \leq thd_{neg}$ {
22:                     **if** data point $S_{b,T}$ is detected as anomalous {
23:                         Increase $C_{agree}$ by one;
24:                         Add data point $S_{b,T}$ to $L_{data}$;
25:                         Add variable $b$ to $L_{var}$; }
26:                     **else** { Increase $C_{disagree}$ by one; }}}}
27:             **if** ($C_{agree} > C_{disagree}$) && ($C_{agree} + C_{disagree} > 1$) {
28:                 Output all data points on $L_{data}$ to be anomalous data points;
29:                 Output all variables on $L_{var}$ to be anomalous variables; }
30:     }} Reset $A$;}

Figure 2: The algorithm of RoLA.

In order to explain how the polling process works, let us look at an example. Assume that the target multivariate time series consists of 5 variables (i.e., $V_1$, $V_2$, ..., and $V_5$). If data point $S_{V_3,2000}$ is detected as anomalous by $LAD_3$, RoLA will calculate the correlation coefficient between variable $V_3$ and every other variable (i.e., $V_1$, $V_2$, $V_4$, and $V_5$) based on Equation 1, where $p$ is a pre-defined integer. Let $a$ denote variable $V_3$, and let $b$ denote every other variable. In this example, $b = V_1, V_2, V_4,$ or $V_5$. If current time point $T$ is less than $p$, RoLA will use all past data points of $a$ and $b$ to compute $E_{a,b}$. Otherwise, it will use the $p$ most recent data points of $a$ and $b$ to calculate $E_{a,b}$.

$$E_{a,b} = \begin{cases} \frac{T(\sum_{z=0}^{T-1} S_{a,z} S_{b,z}) - (\sum_{z=0}^{T-1} S_{a,z})(\sum_{z=0}^{T-1} S_{b,z})}{\sqrt{[T \sum_{z=0}^{T-1} (S_{a,z})^2 - (\sum_{z=0}^{T-1} S_{a,z})^2][T \sum_{z=0}^{T-1} (S_{b,z})^2 - (\sum_{z=0}^{T-1} S_{b,z})^2]}}, & T < p \\ \frac{p(\sum_{z=T-p}^{T-1} S_{a,z} S_{b,z}) - (\sum_{z=T-p}^{T-1} S_{a,z})(\sum_{z=T-p}^{T-1} S_{b,z})}{\sqrt{[p \sum_{z=T-p}^{T-1} (S_{a,z})^2 - (\sum_{z=T-p}^{T-1} S_{a,z})^2][p \sum_{z=T-p}^{T-1} (S_{b,z})^2 - (\sum_{z=T-p}^{T-1} S_{b,z})^2]}}, & T \geq p \end{cases} \quad (1)$$

In Equation 1, $S_{a,z}$ and $S_{b,z}$ are the data points of variables $a$ and $b$ at time point $z$, respectively. The

Pearson correlation coefficient is a statistical measure of the strength and direction of the linear relationship between two variables, and it ranges from −1 to 1. A value of −1 shows a perfect negative correlation, 0 indicates no correlation, and 1 indicates a perfect positive correlation.

In this paper, if the correlation coefficient between $a$ and $b$ (i.e., $E_{a,b}$) is larger than or equal to a positive threshold $thd_{pos}$, it means that $a$ and $b$ have a highly positive correlation. On the other hand, if $E_{a,b}$ is less than or equal to a negative threshold $thd_{neg}$ (where $thd_{neg}$ is negative $thd_{pos}$), it means that $a$ and $b$ have a highly negative correlation. In both cases, RoLA will further check if the current data point of $b$ has been detected as anomalous by its LAD (see line 22 of Figure 2). If the answer is yes, RoLA increases counter $C_{agree}$ by one, add the data point to a list denoted by $L_{data}$, and add $b$ to another list denoted by $L_{var}$ (see lines 22 to 25). Otherwise, RoLA increases counter $C_{disagree}$ by one. The above process is so called polling. Afterwards, if the condition shown on line 27 is evaluated to be true, all data points in $L_{data}$ will be reported as anomalous data points by RoLA, and all variables in $L_{var}$ will be outputted as anomalous variables (see lines 27 to 29).

Let us continue with our previous example. Suppose that $V_4$ and $V_5$ are found highly correlated with $V_3$, but only $V_4$ has its current data point detected as anomalous in addition to the current data point of $V_3$. In this case, $C_{agree}$ will be 2, but $C_{disagree}$ will be 1. Hence, RoLA will report the current data points of $V_3$ and $V_4$ as anomalous data points, and report both $V_3$ and $V_4$ as anomalous variables.

## 4 EXPERIMENT RESULTS

To evaluate RoLA, we conducted an experiment based on the multivariate time series generated by the FerryBox (King et al., 2021; Schroeder et al., 2008) used in the One Ocean Expedition (King et al., 2021; Knoblauch, 2022), which is a circumnavigation of the globe by a Norwegian tall ship called Statsraad Lehmkuhl to share knowledge about the crucial role of the ocean for a sustainable development. The FerryBox uses several sensors to monitor seawater temperature, conductivity, salinity, oxygen concentration, saturation, etc. The dataset consists of the time series of the 12 variables measured every one minute from 2021/10/05 08:49 to 2021/11/29 07:12. However, the time series was not continuous. It could be that the FerryBox was not used all the time.

Therefore, we chose a continuous measurement period from 2021/10/28 00:00 to 2021/10/30 23:59 and used the corresponding multivarite time series to be our target dataset in this experiment. The total number of time points in this datset is 4316, and the following 9 variables are involved: SBE45_Salinity, SBE45_Conductivity, Optode_Concentration, Optode_Saturation, C3_Temperature, Flow_Temperature, Optode_Temperature, C3_Turbidity, and Flow_Flow. Note that SBE45 denotes a temperature monitor called SBE 45 MicroTSG (SEA.BIRD_Scientific, 2023), and it was used by the FerryBox to measure seawater salinity and conductivity. Optode refers to a sensor used to measure seawater temperature, oxygen saturation, and oxygen concentration (AANDERAA, 2023; Schroeder et al., 2008). C3 is a submersible fluorometer that can be configured with up to three or six optical sensors ranging from deep ultraviolet to the infrared spectrum (FONDRIEST_ENVIRONMENTAL, 2023), and it was used to measure seawater temperature and turbidity.

Figure 3 illustrates the time series of these 9 variables. We annotated all obvious anomalies and highlighted each of them in gray. Apparently from Figure 3 that not all the variables are involved in every anomaly. Table 1 lists the details of these anomalies. All the anomalies are collective anomalies since they all last more than one time point, which satisfies the definition of a collective anomaly, i.e., a sequence of data instances that are anomalous compared with the rest of the data (Chandola et al., 2009).

As mentioned earlier, the measurement interval time in the FerryBox dataset is one minute. In order to prove that RoLA can provide immediate response for multivariate time series that are generated on the fly, we reduced the interval time from 1 minute to 10 seconds and injected each 9-dimensional vector of the target multivariate time series from Kafka into RoLA sequentially based on the new interval.

We followed the hyperparameters and parameters used by RePAD2 (Lee and Lin, 2023) to configure the hyperparameters and parameters for each LAD of RoLA. Please see Table 2 for all the values. In other words, users are not required to tune these hyperparameters and parameters for RoLA. In this experiment, RoLA used nine LADs for the nine variables. Each LAD has a simple LSTM network structure with only one hidden layer and ten hidden units, and all LADs adopted Early Stopping (EarlyStopping, 2023) to automatically determine the number of epochs (up to 50) for their online LSTM model training.

RoLA required two parameters to be defined in advance. One is parameter $p$ for correlation coefficient calculation. In this experiment, $p$ was set to 2880, which corresponds to two days for the Fer-ryBox dataset. The other parameter is the positive threshold $thd_{pos}$, which was set to a high value of 0.95 (recall that 1 is the highest value). Note that the negative threshold $thd_{neg}$ does not need to be configured because it equals negative $thd_{pos}$, i.e., $-0.95$.

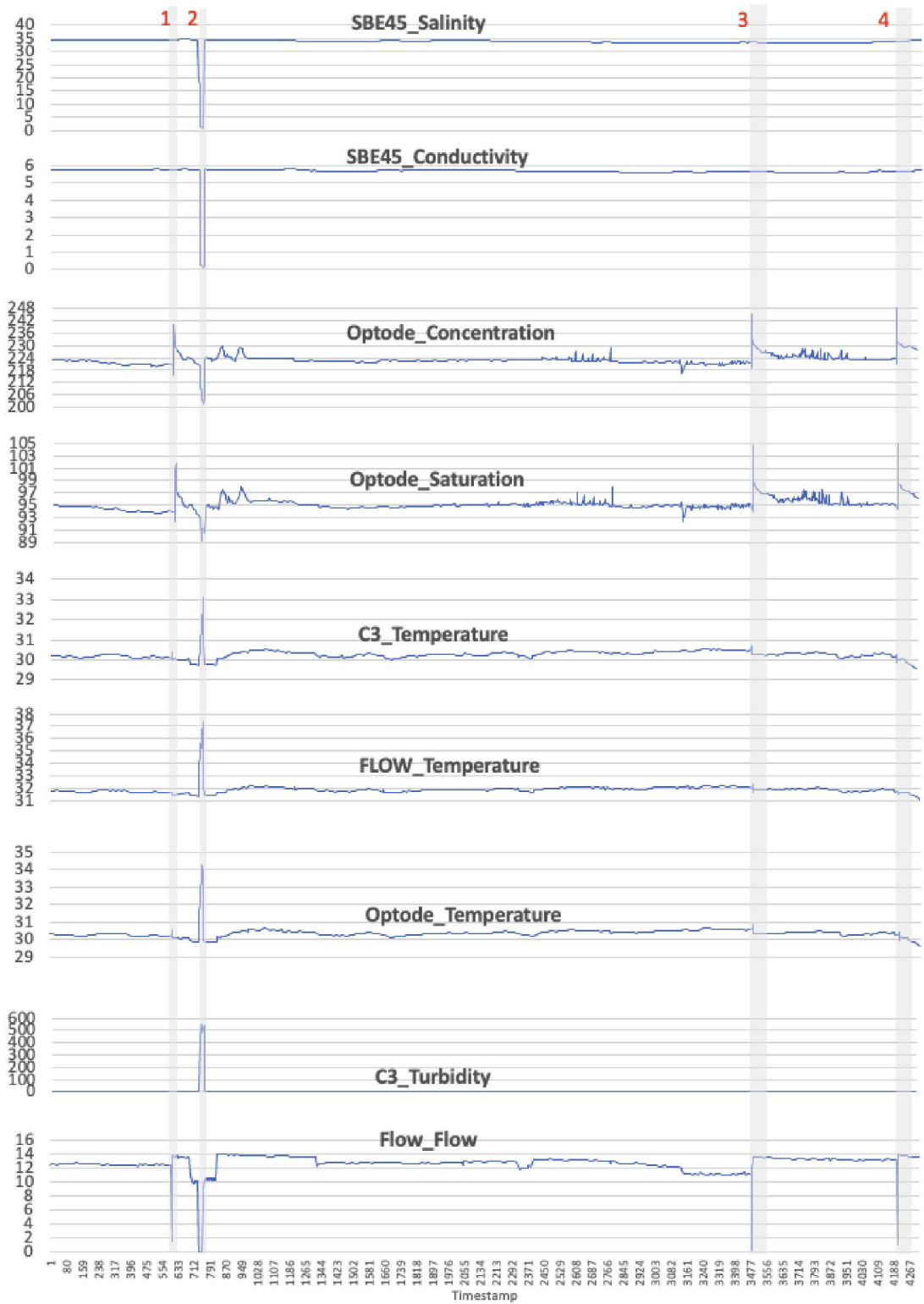

Figure 3: The time series of the nine variables in the target multivariate time series. All anomalies are annotated and highlighted in gray. Note that these anomalies are unknown to RoLA since RoLA is based on unsupervised learning.

Table 1: The details of each annotated anomaly.

| Anomaly No. | Time period | Number of data points | Variables involved |
|---|---|---|---|
| 1 | 2021-10-28 10:05 to 2021-10-28 10:17 | 13 | Optode_Concentration, Optode_Saturation, Flow_Flow |
| 2 | 2021-10-28 12:15 to 2021-10-28 12:39 | 25 | SBE45_Salinity, SBE45_Conductivity, Optode_Concentration, Optode_Saturation, C3_Temperature, Flow_Temperature, Optode_Temperature, C3_Turbidity, Flow_Flow |
| 3 | 2021-10-30 10:08 to 2021-10-30 11:19 | 71 | Optode_Concentration, Optode_Saturation, Flow_Flow |
| 4 | 2021-10-30 22:13 to 2021-10-30 23:29 | 77 | Optode_Concentration, Optode_Saturation, Flow_Flow |

Table 2: The hyperparameter and parameters for each LAD.

| Hyperparameters and parameters | Value |
|---|---|
| The number of hidden layers | 1 |
| The number of hidden units | 10 |
| The number of epochs | 50 |
| Learning rate | 0.005 |
| Activation function | tanh |
| Random seed | 140 |
| The sliding window size $W$ | 1440 |

In this experiment, we implemented RoLA and all the LADs in Deeplearning4j (Deeplearning4j, 2023), which is a programming library written in Java for deep learning. Furthermore, we used Apache Kafka of version 2.13-3.4.0 to inject the target multivariate time series into RoLA sequentially based on the 10-second interval. The entire experiment was performed on a MacBookPro laptop running MacOS Monterey 12.6 with 2.6GHz 6-Core Intel Core i7 and 16GB DDR4 SDRAM. The purpose of using this laptop is to show that RoLA can be very efficient even though it runs on a commodity computer.

As mentioned earlier that RoLA is the first anomaly detection approach that can learn multivariate time series in a completely online manner, and provide real-time and transparent anomaly detection without any labeled dataset, normal dataset, or reference model. According to all the above features, we could not find any similar approach to compare with RoLA. Hence, we chose to compare the joint detection results of RoLA with the individual detection results of LADs.

To easily present the detection performance of RoLA, we illustrated the individual detection results of each LAD on its own variable on the left part of Figure 4 while presenting the joint detection results of RoLA on the right part of Figure 4. Each data point that was detected as anomalous is marked by a small red square. Apparently, when LADs worked on individual variables, each of them generated many false positives, which are the red squares unable to be covered by the gray bars (see the left part of Figure 4). The total number of the false positives is 224.

However, when RoLA was used, it significantly reduced false positives from 224 to 66, implying that combining multiple LADs to jointly detect anomalies is a promising approach. Note that RoLA generated a false positive in the beginning because RoLA found that many variables have a high correlation at that moment. However, this is inevitable since RoLA did not go through any offline training and that it learned only few historical data at that moment. Table 3 lists the detailed detection performance of RoLA. We followed the evaluation method used by (Lee et al., 2020a) to calculate precision, recall (also known as sensitivity), and F-score, which are three widely used metrics for measuring the accuracy of an approach. Note that precision=$TP/(TP+FP)$, recall=$TP/(TP+FN)$, and F-score= $2 \cdot (precision \cdot recall)/(precision+recall)$ where $TP$, $FP$, and $FN$ represent true positive, false positive, and false negative, respectively. As long as any point anomaly occurring at time point $Z$ can be detected within a time period ranging from $Z-K$ to time point $Z+K$ where $K$ was set to 7 following by the suggestion made by (Ren et al., 2019), we say that the anomaly is correctly detected. On the other hand, for any collective anomaly, if it starts at time point $I$ and ends at time point $J$ ($J>I$), and it can be detected within a period between $I-K$ and $J$, we say that the anomaly is correctly detected.

Table 3: The detection performance of RoLA.

| Metric | Value |
|---|---|
| TP | 186 |
| FP | 66 |
| FN | 0 |
| Precision | 0.738 |
| Recall | 1 |
| F-score | 0.849 |

Based on the result listed in Table 3, we can see that RoLA offers satisfactory detection accuracy given the fact that it is a completely unsupervised learning approach and that it uses only online model training without any offline process.

Table 4 details all supportive information outputted by RoLA regarding all detected anomalies. From this table, we can see all variables that were found involved in each anomaly. When Anomalies 1 and 2 just occurred, only few variables were found in-volved. However, when the two anomalies continued (recall that they are both collective anomalies), more variables were found involved because of the obvious changes in their time series. Hence, we showed the largest set of variables that are found involved in each anomaly in Table 4. For Anomaly 1, RoLA found that all involved variables are SBE45_Salinity, Optode_Concentration, Optode_Saturation, C3_Temperature, Flow_Temperature, and Optode_Temperature. However, the real variables involved in this anomaly are only Optode_Concentration, Optode_Saturation and Flow_Flow (see Table 1). We can see that Flow_Flow was not found involved in this anomaly by RoLA even though it had some anomalous data points (see the right part of Figure 4). This is because that RoLA at that moment did not find that all the past data points of Flow_Flow was highly correlated with those of any above-mentioned involved variable.

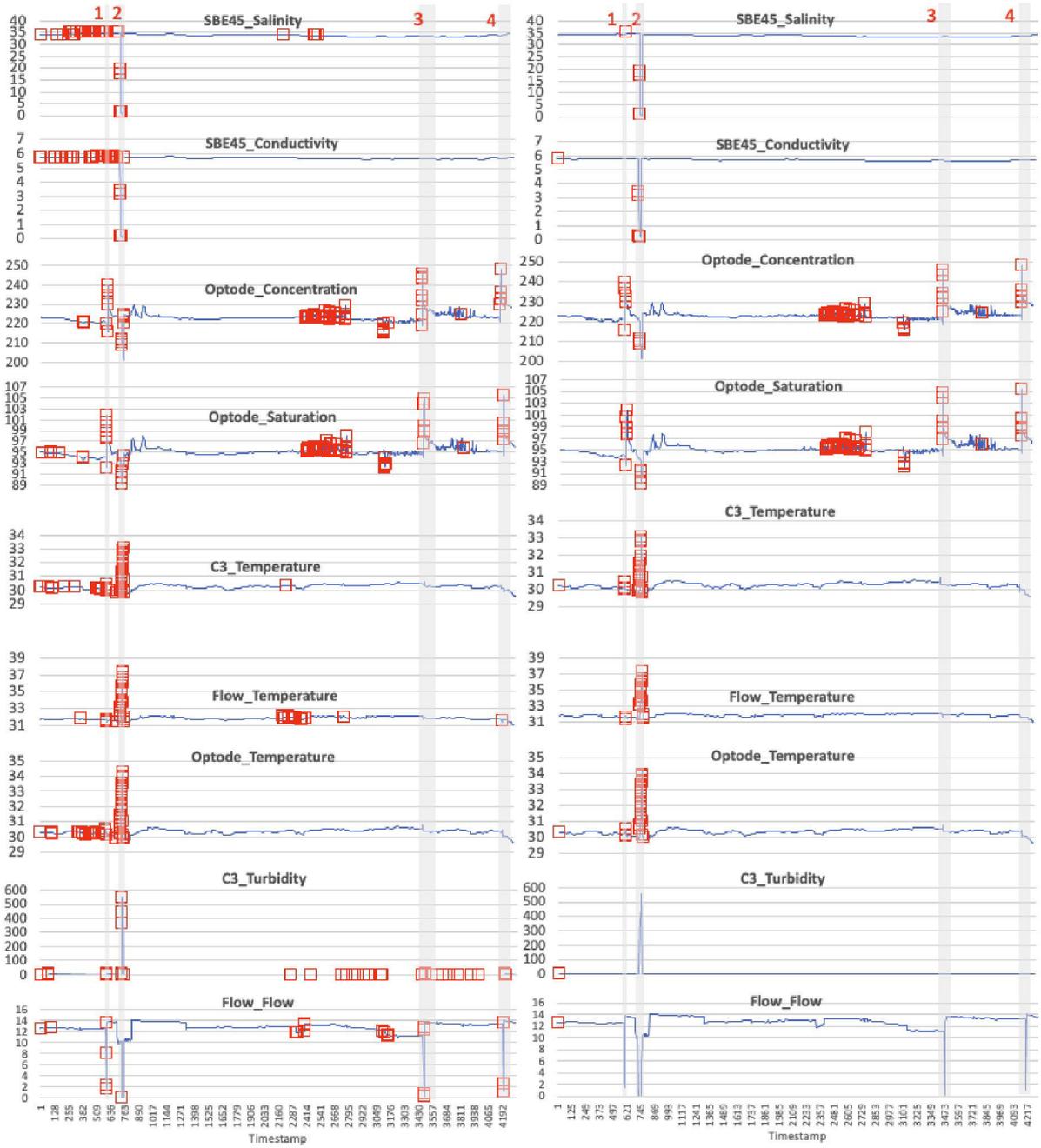

Figure 4: The detection results of the LADs on individual variables (the left part) and the final detection results of RoLA (the right part).

Table 4: The supportive information outputted by RoLA.

| Anomaly No. | All variables that are found involved |
| --- | --- |
| 1 | SBE45_Salinity, Optode_Concentration, Optode_Saturation, C3_Temperature, Flow_Temperature, Optode_Temperature |
| 2 | SBE45_Salinity, SBE45_Conductivity, Optode_Concentration, Optode_Saturation, C3_Temperature, Flow_Temperature, Optode_Temperature |
| 3 | Optode_Concentration, Optode_Saturation |
| 4 | Optode_Concentration, Optode_Saturation |

For Anomaly 2, RoLA found that all involved variables are SBE45_Salinity, SBE45_Conductivity, Optode_Concentration, Optode_Saturation, C3_Temperature, Flow_Temperature, and Optode_Temperature. However, we can see that both C3_Turbidity and Flow_Flow were not found involved since the past data points of these two variables before the occurrence of Anomaly 2 were not highly correlated with any other variable.

In fact, C3_Turbidity and Flow_Flow were not found highly correlated with any other variable in the entire dataset. Nevertheless, all the four anomalies still can be successfully detected by RoLA with other variables. That is why the recall of RoLA is 1.

As for the execution performance of RoLA, we calculated the time required by RoLA to decide whether or not each 9-dimensional vector in the target multivariate time series is anomalous. **The average response time is 0.149 sec with a standard deviation of 0.168 sec**. Due to the fact that the LADs of RoLA need to retrain their LSTM models when they cannot predict well, the standard deviation is slightly higher than the average response time. Recall that we have reduced the interval time of the target multivariate time series from 1 minute to 10 sec in this experiment. The response time result shows that RoLA indeed offers real-time streaming processing and immediate responses even though it runs only on a commodity laptop.

## 5 CONCLUSIONS AND FUTURE WORK

In this paper, we have introduced RoLA for detecting anomalies in a multivariate time series in real time. Different from existing anomaly detection approaches for multivariate time series, RoLA does not need to go through any offline training since it can learn multivariate time series in a completely online manner without requiring any labeled dataset, normal dataset, or reference model. In addition, RoLA does not require users to determine detection thresholds or configure many hyperparameters or parameters.

The most distinct feature of RoLA is its simple and efficient design. By using a LAD to individually learn the time series of each variable, determining highly correlated variables dynamically only when it is needed, and following the majority rule to jointly determine anomalous data points, RoLA can efficiently and effectively detect anomalies and provide useful supportive information for domain experts to further investigate any detected anomaly. The experiment results on the FerryBox dataset show that RoLA offers satisfactory detection performance and instant useful reporting in real time.

In our future work, we plan to improve the detection accuracy of RoLA by reducing false positives. Furthermore, the current version of RoLA will repeatedly calculate correlation coefficients during the occurrence of a collective anomaly. This could be improved to further enhance the performance of RoLA. Another work is to investigate how parameters $p$ and $thd_{pos}$ impact the detection performance of RoLA and to see if there is a way to automatically determine these two parameters.

## ACKNOWLEDGEMENT

The authors want to thank the anonymous reviewers for their reviews and suggestions for this paper.